\documentclass{article}
\usepackage{fancyhdr}

\fancypagestyle{firstpage}{
  \fancyhf{}
  \fancyhead[C]{\scriptsize Second IJCAI AI for Good Symposium in Africa hosted by Deep Learning Indaba}
  
}

\usepackage{ijcai25}

\usepackage{times}
\usepackage{soul}
\usepackage{url}
\usepackage[hidelinks]{hyperref}
\usepackage[utf8]{inputenc}
\usepackage[small]{caption}
\usepackage{graphicx}
\usepackage{amsmath}
\usepackage{amsthm}
\usepackage{booktabs}
\usepackage{algorithm}
\usepackage{algorithmic}
\usepackage[switch]{lineno}

\title{Using Machine Learning to Enhance the Detection of Obfuscated Abusive Words in Swahili: A Focus on Child Safety}

\usepackage{authblk}  

\author{Phyllis Nabangi}
\author{Abdul-Jalil Zakaria}
\author{Jema David Ndibwile}

\affil{College of Engineering, Carnegie Mellon University Africa}

\affil{\texttt{pnabangi@alumni.cmu.edu, azakaria@alumni.cmu.edu, jndibwil@andrew.cmu.edu}}

\begin{document}

\maketitle

\thispagestyle{firstpage}

\begin{abstract}
    The rise of digital technology has dramatically increased the potential for cyberbullying and online abuse, necessitating enhanced measures for detection and prevention, especially among children. This study focuses on detecting abusive obfuscated language in Swahili, a low-resource language that poses unique challenges due to its limited linguistic resources and technological support. Swahili is chosen due to its popularity and being the most widely spoken language in Africa, with over 16 million native speakers and upwards of 100 million speakers in total, spanning regions in East Africa and some parts of the Middle East. We employed machine learning models including Support Vector Machines (SVM), Logistic Regression, and Decision Trees, optimized through rigorous parameter tuning and techniques like Synthetic Minority Over-sampling Technique (SMOTE) to handle data imbalance. Our analysis revealed that, while these models perform well in high-dimensional textual data, our dataset’s small size and imbalance limit our findings’ generalizability. The precision, recall, and F1 scores were thoroughly analyzed, highlighting the nuanced performance of each model in detecting obfuscated language. This research contributes to the broader discourse on ensuring safer online environments for children, advocating for expanded datasets and advanced machine-learning techniques to improve the effectiveness of cyberbullying detection systems. Future work will focus on enhancing data robustness, exploring transfer learning, and integrating multimodal data to create more comprehensive and culturally sensitive detection mechanisms.
\end{abstract}

\section{Introduction}
There is a growing trend of children spending more time online. Globally, a child goes online for the first time every half second \cite{UNICEF2018}. However, this increased digital presence also exposes children to various risks, such as cyberbullying, exposure to hate speech and violent content, and the most alarming threat, online sexual exploitation and abuse \cite{UNICEF2022}. The danger that children face online is supported by a report from the WeProtect Global Alliance, which states that an estimated 750,000 individuals worldwide seek to connect with children online for sexual purposes \cite{weprotect2023}. Furthermore, the UNICEF Protecting Children Online report \cite{UNICEF2022} indicates that around 80\% of children in 25 countries feel threatened by sexual abuse or exploitation online. Furthermore, a recent report on Global Threat Assessment 2023 \cite{weprotect2023} shockingly reveals that 20\% of children in some countries have been subjected to online sexual exploitation and abuse in recent years. These reports further highlight the urgent need for robust measures to safeguard their digital experiences. 

Various forms of abuse have emerged globally, including financial sexual extortion, AI-generated imagery, and child sexual exploitation and abuse \cite{weprotect2}. These abuses are fueled by emerging technologies such as eXtended Reality (XR) \cite{XR}, generative AI \cite{pooyandeh22}, and widespread adoption of end-to-end encryption. Therefore, it is imperative to implement the principles of Safety by Design \cite{weprotect2} and not neglect the prevalent issue of social stigma and taboos in the global south that prevent children from revealing their experiences of online abuse \cite{weprotect3}. 

In investigating and detection of abuse, cultural context is vital and therefore needs to be incorporated. So many things we perceive to be abusive or disrespectful in one culture might have no resemblance or match in other cultures. This cultural variability necessitates that detection systems be both culturally appropriate and adaptable. A study by Ethem, Fazli, et al. on multilingual sentiment analysis demonstrates that models incorporating cultural awareness are more effective at understanding and categorizing sentiments across different languages \cite{chen2018multilingual}.

Also, community-driven efforts and cooperation between academic institutions and technology corporations are critical for closing the resource gap. Masakhane, a grassroots organization focused on NLP for African languages, emphasizes producing and sharing linguistic resources, building open-source tools, and promoting collaborations to improve NLP research in Localizing Low-resource Languages (LRLs) \cite{nekoto2020participatory}. 

Furthermore, with the increasing issues of cyberbullying, the use of Artificial Intelligence (AI) has been suggested to be necessary towards helping achieve realistic mitigation solutions \cite{ptaszynski2018automatic}. One of the popular means to automate the detection of cyberbullying has been to use Machine Learning (ML) and Deep Learning (DL) algorithms which allow the detection of new relevant instances after training them with a sufficient amount of data \cite{eronen2023zero}.
It is against the above background that our research aimed at contributing to the development of an automated solution that complements human efforts in combating cyberbullying amongst children. By providing timely detection, the system aims to foster a safer online environment, protect individuals from cyberbullying incidents, and promote positive digital interactions. Specifically, we aim to improve the detection of obfuscated abusive words in Swahili through the application of machine learning and natural language processing techniques.

The remaining sections of this paper are organized as follows: Section 2 covers the related work on AI-based tools for detecting inappropriate content. Section 3 describes the methods we used and details on how we obtained and processed the dataset, along with the technical specifics of the various ML models employed. The results of our work are presented in Section 4, followed by discussion points in Section 5. Finally, we conclude the study in Section 6 and offer recommendations for future research.

\section{Related Work}

The development of digital technology has expanded the scope for child interaction, education, and socialization, with research increasingly focusing on ensuring their safety online.  Recent studies have revealed various approaches to combat the growing concerns of cyberbullying and online exploitation. Notably, research on the detection of abusive language in low-resource languages has recently drawn much attention \cite{mahmud2023cyberbullying}, \cite{Khairy2023}. 

The Africa Ed-Tech policy brief \cite{Achieng2022} addresses the critical role of technology in protecting children’s privacy on educational platforms across the continent, highlighting the urgent need for strong privacy protections. The brief identifies legislative frameworks and the lack of privacy-preserving measures as a critical concern, an aspect that echoes the need for culturally sensitive machine-learning models that are attuned to regional legislative landscapes and privacy considerations.

The detection of abusive obfuscated language can vary significantly between low-resource languages (LRLs) and high-resources languages (HRLs). Most notably HRLs include English, French, and Chinese benefit from lots of linguistic information (in terms of computer power), extensive annotated corpora, and robust NLP tools. Such benefits ease the building of sophisticated machine-learning models that can detect even malevolent language forms and cyberbullying in disguise. Davidson et al., for example, illustrate how well deep learning models perform when they have access to extensive datasets and pre-trained embeddings in English for hate speech detection \cite{davidson2017automated}. On the other hand, low-resource languages like Kiswahili(Swahili), Zulu, and Amharic have to deal with more disadvantaged settings that come about since they not only require annotated datasets but also linguistic tools which prompt a smaller audience of potential users by virtue of their lesser computation resources available. This uneven distribution poses a significant barrier towards the construction of real systems for detection and prevention of cyberbullying. Studies such as \cite{ruder2019neural} highlight the significant gap in NLP research between HRLs and LRLs, emphasizing the need for more inclusive approaches that consider the unique linguistic and cultural characteristics of LRLs.

Despite these challenges, various interesting ways are being investigated to improve abusive language recognition in LRLs. Transfer learning and multilingual models have demonstrated the potential for exploiting HRL knowledge to improve LRL performance. Conneau et al. (2020) show that cross-lingual language models, such as XLM-R, effectively transfer knowledge across languages, enhancing detection skills in LRLs with limited resources \cite{conneau2020unsupervised}.

While HRLs benefit from advanced resources and tools for detecting obfuscated abusive language, LRLs face significant challenges due to resource scarcity and linguistic diversity. Addressing these challenges requires a multifaceted approach that includes transfer learning, culturally aware models, and community-driven initiatives. As research progresses, the development of more inclusive and effective cyberbullying detection systems would help ensure safer online environments for children across diverse linguistic backgrounds.

It is worth noting that several studies have attempted to contribute towards abusive language detection using various techniques but with a main focus on high-resource languages. These studies are summarized below with their respective gaps as well as the contribution of our research to the body of knowledge.

A study by \cite{Zaccagnino2021} examines using touch gestures as behavioral biometrics to prevent children from accessing inappropriate content. This research raises concerns about the practicality of such measures, given the differences in children's motor skills and the diversity of touch-enabled devices. The further exploration of child safety technology by \cite{Sanchez2019} evaluates the tools used for investigating Child Sexual Abuse Material (CSAM) and highlights limitations in their effectiveness for practitioners, which higlight the necessity for filtering technology capable of detecting and clustering victims' faces and applying age estimation techniques. The general findings indicated that implementing filtering technologies is more important than safe viewing technologies, but practitioners are not well-versed in these technologies. 

The study by Akhter et al., on cyberbullying detection in low-resource languages, such as Urdu highlights the importance of deep-learning models. The study proves that deep-learning models, for example CNNs and RNNs outperform traditional machine-learning methods in terms of accuracy as well as robustness to identify abusive language from social media comments. Deep-learning models are able to capture complex patterns and contextual nuances of obfuscated abusive language, even in difficult conditions due to the scarce linguistic resources. The authors find that larger datasets need to be created and propose extending transfer learning methods can help improve over low-resource languages using multilingual models. Moreover, this work intensifies the importance of leveraging advanced machine-learning algorithms to build robust cyberbullying detection mechanisms across various language backgrounds \cite{Akhter2022}.

Furthermore, \cite{Jonathan2021} explored the use of conventional machine-learning approaches for detecting abusive language in social media comments. Their study primarily focused on Urdu and Roman Urdu languages, employing ML models such as Naive Bayes (NB), Support Vector Machines (SVM), and Logistic Regression. These models were effective in identifying abusive language in standard text formats. However, a significant limitation of their work was its focus on normal text, with limited testing on the models' performance using obfuscated abusive text. This gap emphasizes the need for models that can effectively handle various forms of text manipulation often used to evade detection. Their research highlights the challenges posed by low-resource languages in the domain of abusive language detection emphasizing the importance of utilizing machine-learning models, which have shown promise in handling more complex and nuanced language patterns, including obfuscated text.

Also, \cite{ISAZA2022250} conducted similar research on abusive classification, focusing on online child grooming, by exploring the use of hybrid machine-learning models. Their study incorporated natural language processing (NLP) and semantic analysis, contributing to the discovery of pseudo-intelligent and effective attack classification methods. This led to the design of a prototype that was particularly effective in classifying grooming attacks. They employed NLP techniques such as Word2Vec (W2V) transformation and skip-gram, combined with various ML classification models, to enhance the accuracy and reliability of their system. However, a notable limitation of their work was its focus on normal conversational messages, with a lack of testing on models' performance using obfuscated abusive text and conversations. This oversight underscores the need for models that can effectively handle different forms of text manipulation commonly employed by offenders to evade detection.

Moreover, \cite{article} conducted a comprehensive study aimed at improving the automatic detection of cyberbullying in textual data using both machine learning (ML) and deep learning (DL) algorithms. Their research demonstrated that DL techniques outperformed traditional ML algorithms in terms of accuracy and robustness. Key performance metrics such as precision, recall, and F1-score were highly recommended for evaluating model performance, given the inherently imbalanced nature of cyberbullying-related datasets. The study found that among the various ML models tested, Support Vector Machines (SVM) consistently achieved superior performance compared to other ML algorithms. This highlights the efficacy of SVM in handling the complexities and nuances of cyberbullying detection. However, despite the relative success of SVM, DL models showcased a significant edge, likely due to their ability to capture deeper semantic patterns and contextual information within the text. Furthermore, the study stresses the potential of DL models in advancing the field of cyberbullying detection. By leveraging complex neural networks, DL approaches can better understand and interpret the subtleties of language used in cyberbullying, making them more effective in real-world applications.

Finally, \cite{SimpleClassifier} study aimed at detecting cyber grooming conversations using machine-learning models, specifically Support Vector Machines (SVM) and K-Nearest Neighbors (KNN). Their innovative approach focused on proposing a low-computation cost classification method that leverages various perpetrator attributes to enhance detection performance. The results of their study demonstrated that ML models, particularly SVM and KNN, are highly effective in identifying and classifying abusive texts. By incorporating specific perpetrator attributes into their models, they were able to significantly improve the accuracy and efficiency of the detection process.

It can be observed from all previous research works that none have focused on the detection of obfuscated abusive texts in low-resource languages. While \cite{Njovangwa2021} explored a rule-based approach for detecting obfuscated abusive words, no work, to the best of our knowledge, has applied machine-learning techniques to the detection of Swahili obfuscated abusive words. Consequently, our research aim is to address this gap by leveraging natural language processing and machine-learning techniques to improve the detection of such words. This approach is expected to enhance the development of an automated decision support system capable of more effectively identifying obfuscated abusive words for low-resource languages.

\section{Dataset Description and Detection Models}

The data set used in this study was obtained from a readily available online source. The data set comprises a total of 100 abusive words (50 English and 50 Swahili), 100 abusive texts (50 English and 50 Swahili), 100 obfuscated abusive texts, 100 obfuscated non-abusive texts, and 100 non-abusive texts. For this study, we focused only on the abusive Swahili texts comprising 100 texts. The primary goal is to detect whether a given abusive text is obfuscated. Each text entry is labeled with a binary indicator where `1` denotes an obfuscated abusive text, and `0` denotes a non-obfuscated abusive text.

The detection process involves using machine learning algorithms to classify the texts. The primary challenge lies in distinguishing between obfuscated and non-obfuscated abusive texts. To achieve this, we employ a binary classification model. The model can be mathematically represented as follows:

\begin{equation}
    y = f(x) = 
    \begin{cases} 
      1 & \text{if } x \text{ is obfuscated} \\
      0 & \text{if } x \text{ is not obfuscated}
    \end{cases}
\end{equation}

Where:
\begin{itemize}
    \item $x$ represents the input text.
    \item $y$ represents the binary output indicating whether the text is obfuscated.
    \item $f(x)$ is the classification function implemented by the machine-learning model.
\end{itemize}

We utilized several machine-learning models, including Support Vector Machines (SVM), Logistic Regression, Random Forest, and Decision Trees, optimized through rigorous parameter tuning and techniques like Synthetic Minority Over-sampling Technique (SMOTE) to handle data imbalance. These models were selected for this study due to their well-documented performance on smaller datasets and their interpretability, which is crucial in understanding the classification decisions made in the context of abusive language detection.
The performance of these models was evaluated using precision, recall, and F1-score metrics.

The dataset's imbalanced nature, with a relatively small number of text samples, posed a challenge. Therefore, SMOTE was used to artificially generate more samples in the minority class (obfuscated texts) to ensure a balanced training set. The formula for SMOTE is:

\begin{equation}
    \text{New Sample} = x_i + \lambda \cdot (x_j - x_i)
\end{equation}

Where:
\begin{itemize}
    \item $x_i$ is a sample from the minority class.
    \item $x_j$ is another sample from the minority class.
    \item $\lambda$ is a random number between 0 and 1.
\end{itemize}

This approach helped improve the model's performance by providing a more balanced representation of the classes. The precision, recall, and F1-scores for each model were thoroughly analyzed, highlighting the nuanced performance of each model in detecting obfuscated language.

\subsection{Preprocessing Techniques}
In this research, the preprocessing is applied with the Term Frequency - Inverse Document Frequency (TF-IDF) approach. Results from studies such as \cite{qaiser2018text} and a more recent one by \cite{mambina2022classifying} show that combining TF-IDF with algorithms like Support Vector Machines (SVM) or Logistic Regression provides the model enough information to better detect useful features in text data, thus boosting detection accuracy. It helps converting text data into vectors with numerical values which are useful in detecting obfuscated abusive language as it represents how important and relevant the words appear within texts leading to higher precision/recall rates.

\subsection{Model Training and Evaluation}
Each model was encapsulated within a pipeline that included text vectorization and the classifier itself. Cross-validation was performed using a 5-fold approach to evaluate the models’ performance and ensure robustness.
The TF-IDF vectorization process is defined by the following equations:

\begin{equation}
\text{TF}_{t,d} = \frac{f_{t,d}}{\sum_{t'} f_{t',d}}
\end{equation}

where \(f_{t,d}\) is the frequency of term \(t\) in document \(d\), and the denominator is the sum of the frequencies of all terms in document \(d\).

\begin{equation}
\text{IDF}_{t} = \log \frac{N}{|\{d \in D : t \in d\}|}
\end{equation}

where \(N\) is the total number of documents, and \(|\{d \in D : t \in d\}|\) is the number of documents containing term \(t\).

\begin{equation}
\text{TF-IDF}_{t,d} = \text{TF}_{t,d} \times \text{IDF}_{t}
\end{equation}

The obtained TF-IDF values constitute the features used to train machine learning models. Model evaluation was performed using 5-fold cross-validation on the dataset having been divided into k = 5 subsets. There are 5 folds - in each fold, we use 4 subsets for training and test on the last subset. This process was repeated 5 times, one differentiation holding each time as the validation test set

The accuracy of each model was computed using the formula:

\begin{equation}
\text{Accuracy} = \frac{TP + TN}{TP + TN + FP + FN}
\end{equation}

where \(TP\) is true positives, \(TN\) is true negatives, \(FP\) is false positives, and \(FN\) is false negatives. Other evaluation metrics included precision, recall, and F1-score, defined as follows:

\begin{equation}
\text{Precision} = \frac{TP}{TP + FP}
\end{equation}

\begin{equation}
\text{Recall} = \frac{TP}{TP + FN}
\end{equation}

\begin{equation}
\text{F1-score} = \frac{2 \times \text{Precision} \times \text{Recall}}{\text{Precision} + \text{Recall}}
\end{equation}

These metrics together enabled a holistic judgement of the performance with which all the models picked up on obfuscated abusive language. The conjunction of TF-IDF vectorization and cross-validation gives rise to a very solid, reliable method for model performance evaluation.

\section{Results}
The machine learning models were evaluated using a 5-fold cross-validation method to ensure the robustness and reliability of the results, focusing on accuracy as the main metric. The performance of each model varied, reflecting their different strengths and weaknesses in handling the classification of obfuscated abusive language in Swahili.

The Support Vector Machine (SVM) achieved a mean accuracy of approximately 87\%, demonstrating its effectiveness in high-dimensional spaces and its robustness against overfitting. Logistic regression also performed well with a mean accuracy of around 88\%, indicating its reliability and efficiency in distinguishing between obfuscated and non-obfuscated texts. The Decision Tree Classifier showed the highest mean accuracy of 99\%, suggesting strong performance. In contrast, the Random Forest Classifier showed a lower mean accuracy of approximately 82\%, which was unexpected for an ensemble method. 

The unexpectedly low accuracy of the Random Forest classifier compared to the Decision Tree's high accuracy can be attributed to several factors. The Decision Tree's high accuracy suggests potential overfitting, where the model performs exceptionally well on the training data but poorly on unseen data. Random Forests, designed to reduce overfitting through ensemble learning, may underperform on smaller datasets like the one used in this study, as they require well-tuned parameters and sufficient data to fully leverage their advantages. Additionally, the inherent randomness in Random Forests can sometimes lead to variability in performance, especially if the most informative features are not consistently selected. In contrast, the Decision Tree's simpler structure and use of all features for every split might give it an advantage in this specific context, despite the risk of overfitting.

Figure 1 visualizes these mean cross-validation scores, highlighting the superior performance of the Decision Tree model. Additional evaluation metrics, such as precision, recall, and the F1 score, were considered to provide a comprehensive assessment, as seen in Table 1. These results emphasize the effectiveness of the Decision Tree model, although its high performance requires further investigation to rule out overfitting.

\begin{figure}[h!]
    \centering
    \includegraphics[scale=0.44]
    {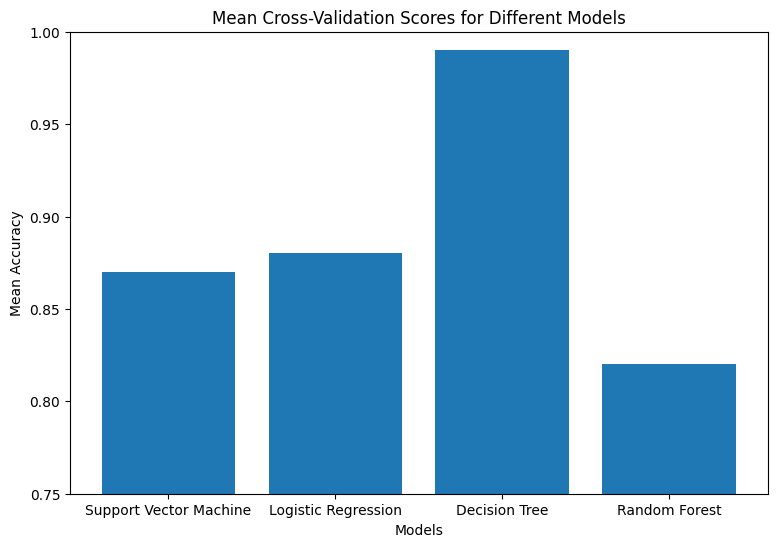}
    \caption{Mean cross-validation accuracy scores for different models}
    \label{fig:model-accuracy}
\end{figure}

\begin{table}[ht]
    \centering 
    \begin{tabular}{llll}
        \toprule
        \textbf{Model} & \textbf{Precision}
          & \textbf{Recall} &\textbf{F1-Score} \\ 
        \midrule
         Logistic Regression & 0.88 & 0.88 & 0.88 \\ 
         Decision Tree & 0.98 & 1.00	& 0.99 \\ 
         Random Forest & 0.79 &	0.86 & 0.83 \\ 
         SVM & 0.86 & 0.88 & 0.87 \\ 
         \bottomrule
    \end{tabular}
    \caption{Model evaluations}
    \label{tab:booktabs}
\end{table}

\section{Discussion}
In this study, we used precision, recall, and F1 scores as performance metrics \cite{metrics}. Each metric offers insight into different aspects of the model performance. The comparison of these metrics allowed us to critically evaluate model performance beyond mere accuracy, ensuring a robust analysis that considers multiple facets of model effectiveness.

The results outlined in the previous section highlight significant differences in the performance metrics of the models evaluated. The Decision Tree model showed high accuracy and precision, its perfect recall score suggests potential overfitting, as it may be too finely tuned to the training data, failing to generalize on unseen data. In contrast, logistic regression and SVM showed more balanced performance, making them suitable for environments where false positives and false negatives carry a similar cost. Future work could further optimize models and explore additional features or advanced techniques to enhance performance, especially for the Random Forest classifier.

\section{Conclusion and Future Work}
\subsection{Conclusion}
This study assessed the effectiveness of various machine-learning models in detecting obfuscated abusive language in Swahili texts. The models included Logistic Regression, Decision Tree, Random Forest, and SVM. Using cross validation, the Decision Tree model exhibited the highest performance with a mean accuracy of 0.99 and precision, recall, and F1-scores of 0.98, 1.00, and 0.99, respectively, indicating excellent accuracy but potential overfitting. Logistic Regression and SVM demonstrated robust performance with mean accuracies of 0.88 and 0.87, respectively, while the Random Forest model showed a lower mean accuracy of 0.82. These results highlight the effectiveness of the Decision Tree model, but prompt the need to address potential overfitting. Additionally, the study's small dataset size may have impacted the generalizability of the results, highlighting the need for larger and more diverse data samples in future research.
In conclusion, this study stresses the potential of machine learning to enhance the detection of abusive obfuscated language in low-resource languages such as Swahili, a critical step toward creating safer digital environments for children. 

\subsection{Future Work}
Despite significant advancements in AI, eXtended Reality (XR), generative AI, and widespread end-to-end encryption, there are enduring challenges in addressing social stigma in regions like the global south, where only a minority of children disclose experiences of online abuse. To bridge these gaps, future research should prioritize expanding datasets and leveraging advanced machine learning techniques to enhance the robustness and cultural sensitivity of cyberbullying detection systems. Integrating multimodal data and exploring innovative approaches such as using touch gestures as behavioral biometrics hold promise, although concerns persist regarding variations in children’s motor skills and device diversity. Addressing these gaps is crucial for fostering safer online environments globally.

Future research endeavors should also focus on mitigating overfitting in models like the Decision Tree by employing techniques such as pruning, while also exploring more sophisticated architectures such as LSTM and BERT to bolster detection capabilities. Enhancing feature extraction methodologies to capture nuanced linguistic patterns and enriching datasets with diverse examples of obfuscated and non-obfuscated texts are imperative steps forward. The development and validation of multilingual models in practical settings will further refine these technologies for real-time deployment, thereby enhancing their practicality and efficacy in detecting obfuscated abusive language. Expanding the scope and diversity of datasets will be pivotal in improving the generalizability and reliability of these models.

Addressing these research gaps through systematic improvements in model training, validation techniques, and dataset enrichment will play a crucial role in advancing the field of cyberbullying detection. By enhancing the effectiveness and inclusivity of detection systems, researchers can contribute significantly to creating safer and more equitable online environments for children worldwide. These efforts highlight the importance of interdisciplinary collaboration and rigorous empirical testing to meet the complex challenges posed by emerging technologies and cultural diversity in online safety.

\section*{Acknowledgments}

The authors would like to express their gratitude to Carnegie Mellon University - Africa for providing funding support through Upanzi Network. This research was conducted as part of a project funded by the Bill and Melinda Gates Foundation. The authors appreciate the financial support that made this work possible.

\bibliographystyle{named}
\bibliography{ijcai25}

\end{document}